\documentclass[preprint, review]{elsarticle}

\usepackage{graphicx}
\usepackage{graphics}
\usepackage{setspace}
\usepackage{amsmath} % assumes amsmath package installed
\usepackage{amssymb,nicefrac}
\usepackage{adjustbox}
\usepackage{amsfonts}
\usepackage{color}
\usepackage{subfigure}
\usepackage[dvips]{epsfig}
\usepackage[dvips]{graphicx}
\usepackage[section]{placeins}
\usepackage{dsfont}
\usepackage{mathrsfs}
\usepackage{sgame}
\usepackage{float}
\usepackage[T1]{fontenc}
\usepackage{aurical}
\usepackage{tikz,pgfplots,verbatim}
\usepackage[linesnumbered,ruled]{algorithm2e}
\usepackage{multirow}
\usepackage{epstopdf}%% The amsthm package provides extended theorem environments
\usepackage{soul}
\usepackage{pdflscape}

\journal{Energy and Buildings}

\definecolor{Yellow}{rgb}{1, 1, 0}
\definecolor{VeryLightGray}{gray}{.90}
\definecolor{LightGray}{gray}{.7}
\definecolor{Gray}{gray}{.50}
\definecolor{DarkGray}{gray}{.3}
\definecolor{VeryDarkGray}{gray}{.10}

\newcommand{\df}{\triangleq}
\newcommand{\STLF}{{\rm{STLF}}}
\newcommand{\AR}{{\rm{AR}}}
\newcommand{\ARMA}{{\rm{ARMA}}}
\newcommand{\ARIMA}{{\rm{ARIMA}}}
\newcommand{\SARIMA}{{\rm{SARIMA}}}
\newcommand{\SARIMAX}{{\rm{SARIMAX}}}
\newcommand{\SARIMAH}{{\rm{SARIMAH}}}
\newcommand{\HW}{{\rm{HW}}}
\newcommand{\HWH}{{\rm{HWH}}}
\newcommand{\HVAC}{{\rm{HVAC}}}
\newcommand{\PAR}{{\rm{PAR}}}
\newcommand{\PARW}{{\rm{PARW}}}

\newcommand{\PARH}{{\rm{PARH}}}
\newcommand{\PAReH}{{\rm{PAReH}}}
\newcommand{\SPR}{{\rm{SPR}}}
\newcommand{\SPRH}{{\rm{SPRH}}}
\newcommand{\SPNN}{{\rm{SPNN}}}
\newcommand{\RLS}{{\rm{RLS}}}
\newcommand{\MLP}{{\rm{MLP}}}
\newcommand{\LSTM}{{\rm{LSTM}}}
\newcommand{\Prophet}{{\rm{Prophet}}}

\newcommand{\RMSE}{{\rm{RMSE}}}
\newcommand{\DD}{{\rm{DD}}}
\newcommand{\ML}{{\rm{ML}}}
\newcommand{\DL}{{\rm{DL}}}
\newcommand{\AI}{{\rm{AI}}}
\newcommand{\MA}{{\rm{MA}}}
\newcommand{\ES}{{\rm{ES}}}
\newcommand{\DT}{{\rm{DT}}}
\newcommand{\SVM}{{\rm{SVM}}}
\newcommand{\PCA}{{\rm{PCA}}}
\newcommand{\LR}{{\rm{LR}}}
\newcommand{\MLR}{{\rm{MLR}}}
\newcommand{\GAM}{{\rm{GAM}}}
\newcommand{\CTMC}{{\rm{CTMC}}}
\newcommand{\DTMC}{{\rm{DTMC}}}
\newcommand{\SGB}{{\rm{SGB}}}
\newcommand{\XGBoost}{{\rm{XGBoost}}}
\newcommand{\RNN}{{\rm{RNN}}}
\newcommand{\ENN}{{\rm{ENN}}}
\newcommand{\ANN}{{\rm{ANN}}}
\newcommand{\BPNN}{{\rm{BPNN}}}
\newcommand{\CNN}{{\rm{CNN}}}
\newcommand{\TNN}{{\rm{TNN}}}
\newcommand{\TCN}{{\rm{TCN}}}

\newcommand{\TFT}{{\rm{TFT}}}
\newcommand{\ReLU}{{\rm{ReLU}}}

\graphicspath{ {./images/} }
\DeclareGraphicsExtensions{.pdf,.jpeg,.png,.jpg}

\newif\iffigures
\figurestrue % uncomment if you want figures
\iffigures
\usepgfplotslibrary{external} 
\begin{document}
\begin{frontmatter}

%% Title, authors and addresses
%% use the tnoteref command within \title for footnotes;
%% use the tnotetext command for theassociated footnote;
%% use the fnref command within \author or \address for footnotes;
%% use the fntext command for theassociated footnote;
%% use the corref command within \author for corresponding author footnotes;
%% use the cortext command for theassociated footnote;
%% use the ead command for the email address,
%% and the form \ead[url] for the home page:
%% \title{Title\tnoteref{label1}}
%% \tnotetext[label1]{}
%% \author{Name\corref{cor1}\fnref{label2}}
%% \ead{email address}
%% \ead[url]{home page}
%% \fntext[label2]{}
%% \cortext[cor1]{}
%% \address{Address\fnref{label3}}
%% \fntext[label3]{}

\title{Hourly Short Term Load Forecasting for Residential Buildings and Energy Communities\tnoteref{t1,t2}}

\tnotetext[t1]{The research reported in this paper has been supported by the Austrian Research Agency FFG through the research project Serve-U (FFG \# 881164). It has also been supported by the Federal Ministry for Climate Action, Environment, Energy, Mobility, Innovation and Technology (BMK), the Federal Ministry for Digital and Economic Affairs (BMDW), and the State of Upper Austria in the frame of SCCH, a center in the COMET - Competence Centers for Excellent Technologies Program managed by Austrian Research Promotion Agency FFG.}
    
%% use optional labels to link authors explicitly to addresses:
%% \author[label1,label2]{}
%% \address[label1]{}
%% \address[label2]{}

%\author{Thomas Grubinger \and Georgios C. Chasparis \and Thomas Natschl\"{a}ger\thanks{Department of Data Analysis Systems, Software Competence Center Hagenberg GmbH, Softwarepark 21, A-4232 Hagenberg, Austria, E-mail: \{thomas.grubinger,georgios.chasparis,thomas.natschlaeger\}@scch.at.} }

\author[scch]{Aleksey V. Kychkin}
\ead{aleksei.kychkin@scch.at}
 %\address{River Valley Technologies, SJP Building, Cotton Hills, Trivandrum, Kerala, India 695014}
   
\author[scch]{Georgios C. Chasparis\corref{cor1}}
\ead{georgios.chasparis@scch.at}

\address[scch]{Department of Data Analysis Systems, Software Competence Center Hagenberg GmbH, Softwarepark 21, A-4232 Hagenberg, Austria}

\cortext[cor1]{Corresponding author}
%\fntext[fn1]{This is the first author footnote.}

\begin{abstract}                % Abstract of not more than 250 words.

Electricity load consumption may be extremely complex in terms of profile patterns, as it depends on a wide range of human factors, and it is often correlated with several exogenous factors, such as the availability of renewable energy and the weather conditions. The first goal of this paper is to investigate the performance of a large selection of different types of forecasting models in predicting the electricity load consumption within the short time horizon of a day or few hours ahead. Such forecasts may be rather useful for the energy management of individual residential buildings or small energy communities. In particular, we introduce persistence models, standard auto-regressive-based machine learning models, and more advanced deep learning models. The second goal of this paper is to introduce two alternative modeling approaches that are simpler in structure while they take into account domain specific knowledge, as compared to the previously mentioned black-box modeling techniques. In particular, we consider the persistence-based auto-regressive model (\PAR{}) and the seasonal persistence-based regressive model (\SPR{}), priorly introduced by the authors. In this paper, we specifically tailor these models to accommodate the generation of hourly forecasts. The introduced models and the induced comparative analysis extend prior work of the authors which was restricted to day-ahead forecasts. We observed a 15-30\% increase in the prediction accuracy of the newly introduced hourly-based forecasting models over existing approaches.
\end{abstract}


\end{frontmatter}

%% \linenumbers
%% main text

\section{Introduction} \label{sec:Introduction}

Widespread implementation of smart energy technologies is increasing user comfort and safety levels as well contributing to the electricity cost reduction. The data generated could be used efficiently to adjust the operation of electrical equipment and reduce the consumption during times of high demand \cite{zhou_bayesian_2016}. Residential buildings connected to energy communities are more likely to provide energy resilience and stability to homeowners \cite{chitsaz_short-term_2015}. Such compact power systems can effectively participate in demand response because of their flexibility and high autonomy. As a consequence, the need to better exploit and balance the energy flexibility potential to/from residential buildings has increased considerably. However, the expected benefit from better exploiting the energy flexibility potential of energy communities is highly dependent on the performance of the forecasting technologies.

\begin{figure*} [t!]
\includegraphics[width=0.98\textwidth]{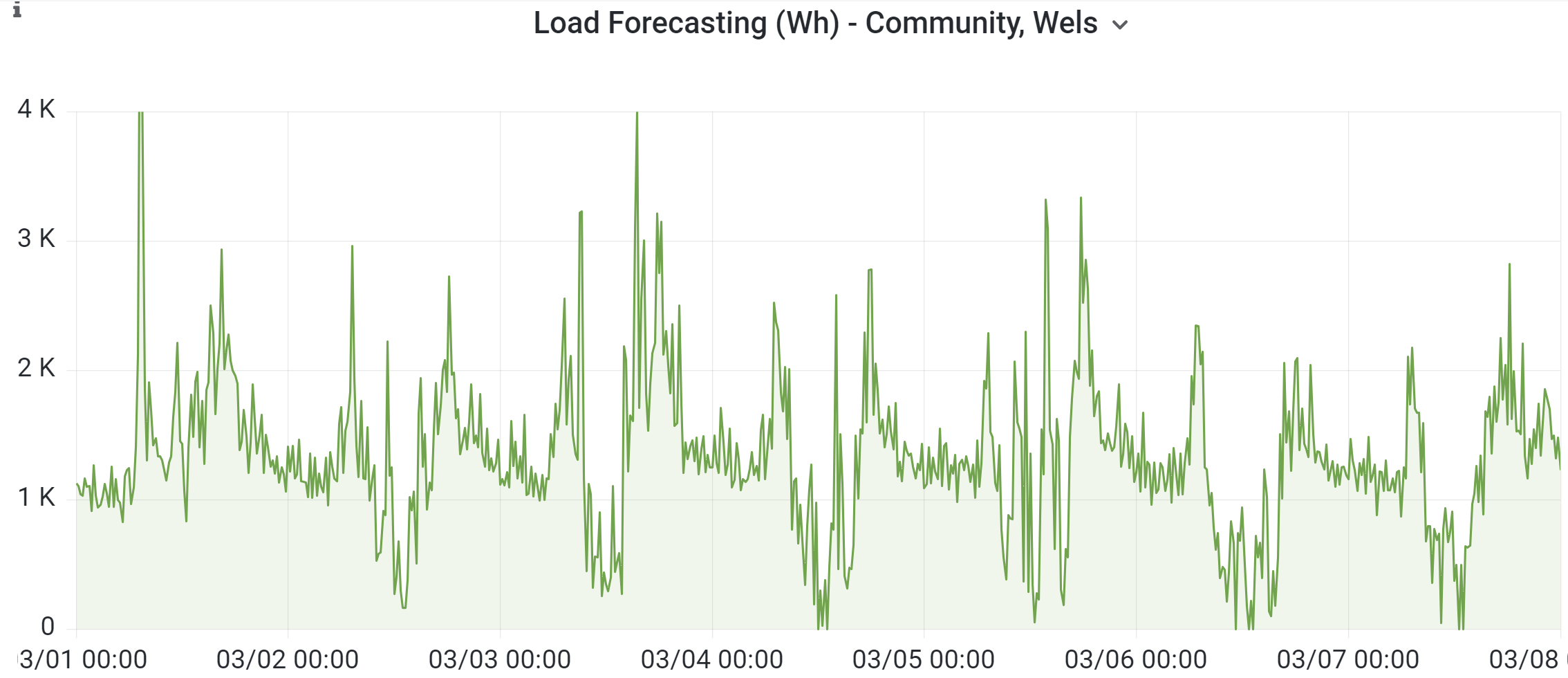}
\caption{Example of the residential Microgrid load over a time period of week (March 2016)}
\label{fig:ElectricityLoadExample}
\end{figure*}

To this end, this paper aims to improve computational efficiency of short-term electricity load forecasts (\STLF) models specialized for either individual residential buildings or small communities. Contrary to prior work that primarily focuses on day-ahead forecasts, our focus is on hourly forecasts. We provide adaptations of persistence-based variations of previously introduced models (namely \PAR{} and \SPR{} introduced in \cite{kychkin_feature_2021}) to capture causal dependencies usually present in users' behavior. Furthermore, we provide a comprehensive comparative analysis with a large selection of alternative commonly used black-box modeling techniques, both in the context of day-ahead and hourly forecasts. The analysis demonstrate the potential of exploiting causal-dependencies in electricity consumption to generate accurate forecasts without increasing computational complexity. This is a considerable advantage compared to computationally intensive black-box modeling techniques.

The remainder of this paper is organized as follows. Section~\ref{sec:RelatedWorkContribution} presents related work and the main contributions of this paper. Section~\ref{sec:ForecastingModelsBackground} presents a collection of different types of forecasting modeling techniques for day-ahead forecasts.  Section~\ref{sec:HourlyModel} introduces adaptations of forecasting techniques to accommodate hourly forecasts. In Section~\ref{sec:Evaluation} we present a comprehensive comparative analysis of the performance of the models, and finally Section~\ref{sec:Conclusions} presents concluding remarks.

%%%%%%%%%%%%%%%%%%%%%%%%%%%%%%%%%%%%%%%%%%%%%%%%%%%%%%
% Related Work and Contribution
%%%%%%%%%%%%%%%%%%%%%%%%%%%%%%%%%%%%%%%%%%%%%%%%%%%%%%
\section{Related Work and Contribution} \label{sec:RelatedWorkContribution}
\subsection{Physics Simulation Models}   \label{sec:PhySimM}

Several research efforts have focused on the accuracy improvements of short-term load forecasting (\STLF) for buildings and communities. Baseline energy consumption calculations are based on the physics simulation principle which provides an estimate of the energy required for a building, primarily for \HVAC{} and light control, e.g., the de.g.ree day (\DD) method \cite{kheiri_split-degree_2023}. Such methods give approximate theoretical values of consumption and do not take into account a mass of factors, including human actions. To overcome this, observational statistics can be used, so that well-known studies include similar day models, expert knowledge, and statistical data analysis techniques including seasonality and exogenous factors like weather \cite{rahman_generalized_1993}.

\subsection{Statistical and Machine Learning Models}   \label{sec:ML}

\subsubsection*{Deterministic Models}

As energy systems of buildings have evolved and user scenarios for managing indoor energy equipment have become more complex, the consumption profile may exhibit irregular patterns, including sudden changes and variations. With the declining performance of classical \STLF{} methods, the researchers focused on the existing statistical predictions through the use of Machine Learning (\ML) techniques \cite{borges_assessing_2013}. Popular models, among the statistical and ML models, include deterministic models, which include persistence models based on historical data from similar days and Moving Average (\MA) \cite{kheiri_split-degree_2023}, Exponential Smoothing (\ES), like Holt-Winters \cite{rahman_generalized_1993}, and adaptive filters for time series, such as the Recursive Least-Squares (\RLS) \cite{khotanzad_adaptive_1995}. These methods also include Decision Trees (\DT), Support Vector Machines (\SVM) \cite{abdoos_short_2015} or Principal Component Analysis (\PCA) \cite{veeramsetty_short_2022} as auxiliary tools to improve the accuracy of other models, such as regression-based or Artificial Intelligence (\AI{}) models \cite{khan_load_2016}.

\subsubsection*{Regression-based models}

Regression-based models, such as Linear Regression (\LR) and Multiple Linear Regression (\MLR), Autoregressive Models (\AR) \cite{hannan_regression_1986}, Box–Jenkins model-based techniques such as \ARMA{}, \ARIMA{} \cite{chen_analysis_1995} models and their variations, often provide high accuracy, especially on small datasets \cite{hora_residential_2022}, \cite{chen_multiscale_nodate}. Such models are univariate in structure and they use only the time series of the energy consumption. Forecasting based on \SARIMA{} provides additionally seasonal effects, and \SARIMAX{} provides exogenous variables processing. Generalized Linear Models and their extension on Generalized Additive Models (\GAM{}) could be multivariate, and they can proceed multiple time series as external features to the load, like weather data or parameters of the building, e.g., the Prophet model from Facebook \cite{stefenon_aggregating_2023}, \cite{riady_multivariate_2023}. 

\subsubsection*{Stochastic Models}

%Existing \STLF{} models are much more diverse in that they are highly dependent on the equipment control modes and the human behavior. As a result, electricity load consumption in residential buildings and small energy communities becomes complex to predict with day-ahead models. 
As the scale of the community increases, there is an increasing influence on the load profile of the largely random intermediate energy flows, which is generally characterized by a high degree of uncertainty and the human factor. The stochastic nature of the residential microgrid electricity consumption eventually causes deviations from expected demand. Among the known statistical and ML methods used for time series forecasting, there is also a group of methods based on the stochastic principle, the essence of which is to reduce the uncertainty of the description of the Load state at a new point in time on a probabilistic basis. Such methods, like Markov Chains (Continuous Time Markov Chains – \CTMC and Discrete Time Markov Chains - \DTMC) operate well with random factors of profiles, such as the human factor \cite{duan_fast_2019}, \cite{hutchison_markovian_2012}, \cite{ruan_stochastic_2017}. Furthermore, Stochastic Gradient Boosting (\SGB) can be used for \STLF{} problems to sequentially fit a simple parameterized function or aggregate (ensemble) of several predictive models \cite{nassif_short_2016}. \SGB{} is also known as an Extreme Boosting (\XGBoost).

The ensemble mechanism is universal and can be applied to any basic statistical, ML or AI and Deep Learning models. Besides \XGBoost, there are many other ways of adding models to an ensemble based on Gradient Boosting core, such as Light Gradient Boosting (LightGBM) or CatBoost. Their essence is to construct models in such a way that they are maximally correlated with the negative gradient of the loss function of the whole ensemble.

\subsection{\AI{}-powered and Deep Learning Models}   \label{sec:AI}

Nowadays \STLF{} is powered by \AI{} and tackled through Deep Learning (\DL) methodologies that are widely implemented both at the low or building level and the high or Transmission System Operator (TSO) level \cite{behm_how_2020}. Multilayer Perceptron (\MLP{}) \cite{khotanzad_adaptive_1995}, \cite{behm_how_2020}, Recurrent Neural Networks (\RNN) \cite{vanting_evaluation_2022} such as Elman Neural Network (\ENN) \cite{xie_short-term_2020} and Long Short-Term Memory (\LSTM) \cite{hochreiter_long_1997}, \cite{wang_electricity_2023}, \BPNN \cite{singh_comparative_2014} were the first models for time series prediction, but other architectures using Convolutional Neural Networks (\CNN) \cite{aouad_cnn-sequence--sequence_2022} and Transformers (\TNN) \cite{lheureux_transformer-based_2022} are also increasingly being used. 

% \CNN{} models are neural networks designed to analyze temporal data, where convolution layer blocks are used to process fragments of a time series and allow the detection of dependencies between different variables, trends and cycles. One kind of architecture based on this is the Temporal Convolution Network (\TCN) / DeepTCN, TimesNet \cite{zuo_ensemble_2023}, \cite{zheng_tcn-gat_2023} and Deep Mind's WaveNet \cite{dorado_rueda_short-term_2021}. Transformers have a neural network structure that uses the Attention mechanism to allow for capturing the complex dynamic nonlinear sequence dependencies on the long sequence input. Multi-Head Attention (\MHA) is a technique embedded in the Transformers structure that allows the model to focus on multiple aspects of the input sequence simultaneously, helping to generate more accurate load predictions. \MHA{} is a key factor in Transformers models reaching new heights of accuracy in load profile processing tasks such as time series, including time information, such as that done in the Time Augmented Transformer (\TAT) \cite{zhang_short-term_2022}. The Temporal Fusion Transformers (\TFT) architecture, which was originally developed to create interval forecasts of multivariate time series, combines the \LSTM{} blocks and the Transformers engine to make it one of the highest performing architectures currently available \cite{wang_transformer-based_2022}.
%
These models try to capture hidden energy consumption dependencies within daily sequences of load values, climatic and meteorological data, and calendar data. Of course, the forecasting accuracy can be increased by incorporating the human behavior. However, the research of private user activities carries certain security risks, including personalization of the detected patterns and the identification of temporal dependencies of presence \cite{taik_electrical_2020}. % This leads to a contradiction between the requirements of providing depth of personification and the required level of forecasting quality based on and identification of user load profile patterns. To overcome this, personalized models can be trained for each building individually using Federated Learning tools.

\subsection{Other Data Driven Models}   \label{sec:Data}

Processing building energy consumption data using other data-driven models, including Expert Systems, Fuzzy Logic \cite{debnath_forecasting_2018}, Fourier \cite{yukseltan_hourly_2020} and Wavelet Transform \cite{li_short-term_2015}, holds great potential for discovering systematic patterns hidden from ML or AI powered models. For example, the paper \cite{pflugradt_synthesizing_2017} introduced an approach for energy load simulation of the behavior of the residents in the household as independent, desire-driven agents. %Developed model can simulate office workers, unemployment, retirees or factory shift workers, or  multifamily houses in detail, including details such as illness periods, personal hobbies and device ownership. 
The division of the day into zones and the forecasting of individual parts, including the forecasting separately of different days of the week, is discussed by \cite{yukseltan_hourly_2020}. An hourly forecasting methodology was proposed in which the hourly prediction by a Fourier series expansion is updated by feeding back the prediction error at the current hour for the forecast at the next hour multiplied by \AR{} model-based optimized scaling factor.

\subsection{Overview}

In summary, methods from the first group (i.e., \emph{Physics Simulation Models}) try to analytically describe the building load based on equations that is very sensitive to completeness and correctness of variables and initial conditions; many factors affecting consumption are hardly considered. The second group (i.e., \emph{Statistical and Machine Learning Models}) process historical data, which in many respects more correctly describes the forecast consumption for a day ahead, taking into account the uncertainty in the influence of factors, and also allows to determine the trend, seasonal dependencies, and systematic components at the daily level. Such models smooth the data and as a result cannot predict well the electricity load with irregular peaks. Such cases are better addressed by models of the third group (i.e., \emph{{\AI} and {\DL} Models}). These models help to identify nonlinear dependencies in the energy data and hidden patterns in the day consumption, but usually require larger training times and data sets. Finally, models of the fourth group (i.e., \emph{Other Data Driven Models}) although having great potential in detecting patterns in human behavior and its effect on the electricity load, are not sufficiently robust and require accurate parameterization. An overview of the hourly-based \STLF{} methods presented above is also provided as a list in Table~\ref{tb:ResultsLiterature}. 

\begin{table*}[h!]
\begin{center}
\caption{Overview of hourly-based short term load forecasting applications on buildings, communities and areas} \label{tb:ResultsLiterature}
\footnotesize
\begin{tabular}{ccccc}
Consumer & Model class & Algorithm & Year & Ref. \\ \hline

Single Building	& Physics Simulation & \DD & 2023 & \cite{kheiri_split-degree_2023} \\ \hline

Municipal utility &	Statistical	& LR & 1993 & \cite{rahman_generalized_1993} \\ \hline

Single Building	& Machine Learning & \AR, \SVM & 2013 & \cite{borges_assessing_2013} \\ \hline

Municipal utility & Statistical & \ES, \MLR & 1991 & \cite{rahman_priority_1991} \\ \hline

Single Building	& Machine Learning & \LR, \ARIMA, \ANN & 2022 & \cite{hora_residential_2022} \\ \hline

Industrial Building	& Machine Learning & Prophet & 2023 & \cite{riady_multivariate_2023} \\ \hline

Single Building	& Machine Learning & \CTMC, \DTMC & 2012 & \cite{hutchison_markovian_2012} \\ \hline

Commercial \tabularnewline / residential buildings & Machine Learning & \SGB & 2016 & \cite{nassif_short_2016} \\ \hline

Country	& {\AI} and Deep Learning & \MLP as \ANN & 2020 & \cite{behm_how_2020} \\ \hline

Municipal utility & {\AI} and Deep Learning & \MLP, \RLS & 1995 & \cite{szmit_implementation_2012} \\ \hline

Residential areas & {\AI} and Deep Learning & \RNN{} & 2022 & \cite{vanting_evaluation_2022} \\ \hline

Residential areas & {\AI} and Deep Learning & \LSTM{}, seq2seq & 2023 & \cite{wang_electricity_2023} \\ \hline

Municipal utility & {\AI} and Deep Learning & \ENN{}, \BPNN{} & 2014 & \cite{singh_comparative_2014} \\ \hline

Single Building	& {\AI} and Deep Learning & \CNN{}, seq2seq & 2022 & \cite{aouad_cnn-sequence--sequence_2022} \\ \hline

Residential areas & {\AI} and Deep Learning & \TCN{}, TimesNet & 2023 & \cite{zuo_ensemble_2023} \\ \hline

Residential areas & {\AI} and Deep Learning & \TNN{} & 2022 & \cite{zhang_short-term_2022} \\ \hline

University Buildings & {\AI} and Deep Learning & \TFT{} & 2022 & \cite{wang_transformer-based_2022} \\ \hline

Single Building & {\AI} and Deep Learning & \LSTM{} & 2020 & \cite{taik_electrical_2020} \\ \hline

Single Building	& Other & Basic Behavior Model & 2017 & \cite{pflugradt_synthesizing_2017} \\ \hline

Residential areas & Other & Fourier Transform, \AR{} & 2020 & \cite{yukseltan_hourly_2020} \\ \hline 

\end{tabular}
\end{center}
\end{table*}

\subsection{Contributions}   \label{sec:Contribution}

Demand response in individual residential buildings or small energy communities requires a high forecasting accuracy of the electricity load consumption. This need is even more evident when demand response needs to be performed over the next hours. In this case, \STLF{} need to also operate in an hourly basis which brings up new challenges with respect to both computational efficiency and incorporating trends in user patterns/behavior. These challenges are highly realized when trying to incorporate standard black-box modeling techniques into the framework of hourly-based \STLF{}. 

In this paper, we provide a thorough investigation of several types of forecasting models, and we perform an extensive comparative analysis using real-world electricity consumption data from a small energy community in Wels, Upper Austria. Moreover, we extend two types of persistence-based regression models introduced by the authors in \cite{kychkin_feature_2021} (originally designed for day-ahead forecasts) to also accommodate hourly-based forecasts. Such simpler models provide computational efficiency in the generation of hourly forecasts due to their simple structure, while they achieve a significant improvement (in the range of 15-30\%) over several standard black-box modeling techniques including state-of-the-art AI-based forecasting models.

This paper is also a contribution over its earlier version appeared in \cite{kychkin_feature_2021}, since it provides a more thorough comparison with advanced AI-powered forecasting techniques, including the \LSTM{} model, and Facebook`s Prophet  model. Due to the presented extensive comparative analysis with state-of-the-art AI-based forecasting methods, this paper also serves as a review paper on the most recent \STLF{} methodologies.

\section{Forecasting Models Background}
\label{sec:ForecastingModelsBackground}

In this section, we provide the necessary background for setting up \STLF{} forecasting models. We present several types of forecasting models that will later be used as a basis for hourly-based electricity load forecasting. We will classify these models into different categories depending on the information or types of features used as well as their modelling structure models (e.g., statistical, autoregressive, deep-learning based, etc.). 

\subsection{Notation}   \label{sec:Notation}

The following notation will be used. Each day, indexed by $d$, is divided into a set of time intervals, indexed by $t$. We assume that each time interval has size  $T$ in $min$ (usually $T=15min$). The total number of time intervals within one day will be denoted by $K\df 1440/T$. The electricity load at time index $t$ and day $d$ will be denoted by $y_d(t)$, and the forecasting result given by model $\times$ will be denoted by  $\hat{y}_d^{\times}(t)$. Forecasts are updated at specific time instances during the day with a period $T_h$ (e.g., $T_h=4h$). In certain cases, we will be using the notation $K_h\df T_h/T$ to denote the number of intervals between two forecast updates, e.g., when $T_h=4h$ then $K_h=16$. 

% \subsection{Framework}  \label{sec:Framework}

% In this section, we present the basic framework of day-ahead \STLF{} experimental setup, Fig. \ref{fig:STLF_schema}. 
% %
% \begin{figure}[th!]
%     \centering
%     \includegraphics[scale=0.70]{Hourly based STLF_2.png}
%     \caption{Schema of STLF model architecture implementation}
%     \label{fig:STLF_schema}
% \end{figure}
% %
% The scheme is designed to take into account existing knowledge in building profile research, and to validate new components based on \AI{} and Deep Learning. The given models in blocks are connected to each other in a specific way, some of the models are used for validation needs in the form of a "black-box". The diagram shows the main data flows, so that the input $y$ denotes the vector of load measurements for the Microgrid, $W$ denotes the vector of weather data; the intermediate vectors $y_i$ denote the prediction results of the individual models; finally, the output matrix $Y$ contains the prediction vectors of all models.

% From a mathematical point of view, we have to introduce refinements. Here we introduce a few notations that will be used further for describing {\STLF} models: 

\subsection{Persistence models} \label{sec:PersistenceModels}

Persistence models constitute computationally efficient forecasting models, which are based on averaging the load consumption over previous days and for the interval of interest. Such models may establish baseline predictions that can be used for benchmarking with more advanced forecasting models. The paper considers two types of persistence models: the first one relies on $N$ previous calendar days and the second one relies on $N$ previous same days, i.e., days with the same position in the week. 

More specifically, the \emph{\textbf{$N$-day persistence model}} (or briefly $N$-day) is defined as follows 
\begin{equation}    \label{eq:NdayPersistenceModel}
\hat{y}_d^{\rm PM} (t) = \frac{1}{N}\sum_{i=d-N}^{d-1}y_{i}(t), 
\end{equation}
where $N$ is the length of the history in days used for generating the forecast. The second one, namely \emph{\textbf{$N$-same-day persistence model}} (or briefly $N$-same-day) computes forecasts as follows 
\begin{equation}    \label{eq:NsamedayPersistenceModel}
\hat{y}_d^{\rm PM}(t) = \frac{1}{N}\sum_{i=1}^{N}y_{d-7i}(t).
\end{equation}
% where $N$ again here define the length of the history in days (however the days considered have the same position within one week).

Each of these models builds an average load profile over several days, considering the mean energy consumption for each time interval $t$. Taking into account the possibly high granularity of the measurements, persistence models are able to describe daily patterns, provided that they are repeated in the analyzed sequence at the same time points during the day. We also provide ensemble samples of \emph{\textbf{N-day persistence model}} via \XGBoost{} to force nonlinear dependencies detection in a load data.

\subsection{Triple exponential smoothing Holt-Winters (\HW) model}  \label{sec:TripleExponentialSmoothing}

The \HW{} model, also known as \emph{triple-exponential smoothing}, cf.,~\cite{kalekar_time_2004}, is a season-based algorithm that tries to capture repeated fluctuations within a fixed period of the time series, \cite{szmit_implementation_2012}. In our case the energy load predictions could be estimated by \HW{} model with daily seasonality as follows: 
\begin{subequations}    \label{eq:HW}
\begin{equation}    
\hat{y}^{\rm HW}_d(t) = \ell(t-k)+k\cdot \tau(t-k)+ \sigma(t-L), 
\end{equation}
where 
\begin{eqnarray}
\ell(t) & \df & \alpha\cdot \big(y(t)-\sigma(t-L)\big)+ (1-\alpha)\cdot \big(\ell(t-1)+\tau(t-1)\big), \label{eq:HW:LevelComponent} \\  	 
\tau(t) & \df & \beta \cdot \big(\ell(t)-\ell(t-1)\big)+(1-\beta) \cdot \tau(t-1), \label{eq:HW:TrendComponent} \\
\sigma(t) & \df & \gamma \cdot \big(y(t)-\ell(t-1) - \tau(t-1)\big)+(1-\gamma)\cdot \sigma(t-L) \label{eq:HW:SeasonComponent}
\end{eqnarray}
\end{subequations}
are the \emph{level component}, \emph{trend component} and \emph{seasonal component}, respectively. 

We parameterize this model to better capture the repeated daily patterns in the electricity load consumption. To this end, the \emph{level component} that represents the corresponding load at the previous time interval, i.e., $k=1$ time intervals. This also determines the setup of the \emph{trend component}, which will capture the trend in the consumption load over two consecutive intervals. The \emph{seasonal component} is determined by the season length which is set to $L=96$, i.e., to one day. The learned parameters are the level smoothing factor $\alpha\in(0,1)$, the trend smoothing factor $\beta\in(0,1)$ and the seasonal smoothing factor $\gamma\in(0,1)$.   
% We have used the following notation: $k$ is the forecasting range $k$=96 (which  refers to the granularity of sensor data within one day), $L$ is the length of the season (here $L=96$), $\alpha\in(0,1)$ is the data smoothing factor, $\beta\in(0,1)$ is the trend smoothing factor, and $\gamma\in(0,1)$ is the seasonal change smoothing factor, that tries to capture daily patterns. The \emph{level component} (or \emph{deseasonalized component}) estimates a baseline load, which is updated by deseasonalizing the data so that only the trend and level component enter to its updating process. The \emph{trend component} is simply the smoothed difference between two successive estimates of the level or deseasonalized component. Finally, the \emph{seasonal component} is a combination of the most recently observed seasonal factor, $y(t)-\ell(t-1)-\tau(t-1)$ (which is computed by the observation when removing the level and trend component), and the previous best seasonal factor estimate for this period, $\sigma(t-L)$. Thus, 
Overall the \HW{} model with daily seasonality is able to observe a combination of both daily and hourly temporal dependencies. 

\subsection{Seasonal autoregressive integrated moving average (\SARIMA) model}

A variation of the \ARIMA{} (Auto-Regressive Integrated Moving Average) model \cite{hamilton_time_1994}, namely \SARIMA{} can be used instead to also track daily seasonality. This model uses three main parameters $(p,r,q)$ as the non-seasonal parameters, where $p$ is the auto-regressive order, which allows to incorporate previous values of the time series; $r$ is the order of integration, which allows to incorporate previous differences of the time series; and $q$ is the order of the moving average, which allows for setting the model error as a linear combination of previously observed error values.

In addition, \SARIMA{} also introduces parameters $(P,R,Q)$ which are defined similarly to $(p,r,q)$, but apply instead to the seasonal component of the time series. Finally, parameter $S$ describes the period of the season in the time series (96 if the season corresponds to one day, where 96 refers to the granularity of sensor data within one day). 

In particular, the \SARIMA{} model can be defined as follows
\begin{equation}    \label{eq:SARIMA}
    \phi(z)\Phi(z^m)(1-z)^d(1-z^m)^D y(t) = \theta(z)\Theta(z^m)\epsilon(t)
\end{equation}
where $z$ corresponds to the one-step delay operator, and $\phi$, $\Phi$, $\theta$ and $\Theta$ correspond to polynomials of the form 
\begin{equation*}
\phi(z) \df 1-\sum_{j=1}^{p}\phi_jz^j\,, \theta(z) \df 1 - \sum_{j=1}^{q}\Phi_jz^j \,,
\Phi(z) \df 1-\sum_{j=1}^{P}\phi_jz^j \,, \Theta(z) \df  1 - \sum_{j=1}^{Q}\Phi_jz^j
\end{equation*}
Compared to the standard \ARIMA{} model, we have the seasonal auto-regressive terms $\Phi(z^{m})$ of order $P$, difference terms of order $D$, and the moving average terms $\Theta(z^m)$ of order $Q$, where the seasonal period is assumed of $m$ steps.

The selection of the parameters $(p,r,q)(P,R,Q)S$ was based upon the recommendations presented in \cite{akpinar_year_2016}. Therein, it is recommended that the conditions $r + R \leq 2$, $P + Q \leq 2$ should be satisfied. In order to compute the most appropriate set of parameters, we first created an enumeration of the model parameters, which were then compared by using the Akaike information criterion (AIC). This process led to the following \SARIMA{} parameters: (1,1,1)(1,1,1)96. The previously described model is set to provide either hourly or daily forecasts by simply adjusting the forecast horizon.

\subsection{Generalized additive model based on Facebook`s Prophet algorithm}
\label{sec:Prophet}

A detailed description of the methodology implemented in Prophet model from Facebook can be found in the articles \cite{stefenon_aggregating_2023}, \cite{riady_multivariate_2023}, \cite{bashir_short_2022}. This methodology is based on the Human-in-the-Loop modelling and provide fitting additive regression models (Generalized Linear Models and their extension on Generalized Additive Models, \GAM) of the following form:
\begin{equation}    \label{eq:Prophet}
\hat{y}_d^{\rm Prophet}(t)= g(t) + s(t) + h(t) + w_t
\end{equation}
where $g(t)$ captures non-periodic changes in the time-series (trend), $s(t)$ captures seasonal fluctuations (daily, weekly, etc.), $h(t)$ captures the effects of holidays and other significant calendar events, and $w_t$ corresponds to a normally distributed random disturbance. 

In particular, to approximate the trend of the series, the model utilizes either a piecewise linear regression or a saturating growth model. In this paper, we will consider the second option described by 
\begin{equation}    \label{eq:GAM_g1}
    g(t) = \frac{C}{1 + e^{-k(t-m)}}
\end{equation}
where $C$ denotes the carrying capacity of the model, $k$ represents the growth rate,  and $m$ is the data offset. To approximate the annual, weekly or daily seasonality, we use partial sums of a Fourier series as follows
\begin{equation}    \label{eq:GAM_s}
s(t) = \sum_{n=1}^{N} \left\{a_n cos\left(\frac {2\pi nt}{P}\right) + b_n sin\left(\frac {2\pi nt}{P}\right)\right\} 
\end{equation}
where $a_n$ and $b_n$ are the learning parameters, $P$ is the regular period of the load sequence, and $N$ is the period length used in the model. As expected, this model finds a natural fit to the problem of forecasting day-ahead load consumption, since in this case $s(t)$ can be used to describe daily seasonality/patterns.

% The effect of the ``holidays'' (e.g. official holidays and weekends as well as other special events) is captured by indicator variables for each independent model separately as follows:
% \begin{equation}    \label{eq:GAM_h}
% h(t) = Z(t)\kappa \,, \quad \kappa \sim {\rm Normal}(0,\nu ^2)
% \end{equation}  
% where $Z(t) = [{\bf 1}_{t\in D_1},\ldots,{\bf 1}_{t\in D_L}]$, with $\bf{1}$ denotes the index function. The term $\kappa$ shows the impact of the time period before and after a particular holiday $i$ with $D_i$ window period. The value of the standard deviation $\nu$ determines the impact of the holiday on the model in this way that a larger value of $\nu$ indicates that the holiday has a greater impact, while a smaller value indicates a smaller impact.

\subsection{Long Short-Term Memory Model (\LSTM)}  \label{sec:LSTM}

Observations of electricity consumption in residential buildings and their communities have shown that the statistical characteristics from week to week vary within narrow limits, in other words, the patterns of weekly variations in electricity consumption are maintained. However, there are huge variations within hourly intervals due to the user behavior. For these consumption cases, an in-depth analysis of intra-daily sequences is required, which can be provided by the \LSTM{} artificial neural network due to its ability to learn long-term correlations, efficiently process time series data in short-term sequences, automatically detect and learn patterns of complex sequences, and adapt to changing input data \cite{hochreiter_long_1997}, \cite{wang_electricity_2023}.

The \LSTM{} architecture consists of three main components: the ``forget gate'' with values $f_t$, the ``input gate'' with $i_t$ and the ``output gate'' with $o_t$. The learning phase implies that the first layer data is processed by using a sigma function. The non-stationary source data is excluded, and the remaining data is moved to the next step. The data is sorted using the following expression: $f_t = \sigma (W_f[h_{t-1} , x_t] + b_f),$ where $W_f$ is the input coefficient, $h_t$ is the output vector as a hidden state, and $b_f$ is the input layer threshold. The next process trains the selected input data to determine the prediction indicator and determines their acceptable values. The sigma and tangent functions are used in this process: $i_t = \sigma (W_i[h_{t-1} , x_t] + b_i),$ and a cell state $c_t = {\rm tanh} (W_c[h_{t-1} , x_t] + b_c),$ where $x_t$ is the input data, and $b_i$, $b_c$ are the neuron thresholds. Based on the selected data, the neurons in the output layer are identified by systematically combining sigma and tangent functions. That is, the prediction result is determined in this process: $o_t = \sigma (W_o[h_{t-1} , x_t] + b_o),$ and finally $h_t = o_t {\rm tanh}(c_t),$ where $o_t$ is the initial data selected using the sigma function, and $h_t$ is the neuron network output layer.

In this paper, we set the parameters of the LSTM architecture as follows (assuming a granularity of $15min$ in the measurement data): the number of input layer neurons is $96\times 10$ (i.e., historical data from 10 previous days are used as an input), the number of neurons of the first hidden recursive layer is $96\times{5}$, the number of neurons of the second hidden recursive layer is $96\times{3}$, the number of neurons of the output layer is $96$. Such configuration gives LSTM the ability to capture daily and hourly patterns within existing sequences of daily load.

\subsection{Persistence-based autoregressive (\PAR) model}  \label{sec:PARmodel}

As was discussed in Section~\ref{sec:PersistenceModels}, the persistence models can capture daily temporal dependencies in the sequence of load profiles, while auto-regressive models are able to capture hourly temporal dependencies. The \PAR{} model for daily forecasts, introduced in \cite{kychkin_feature_2021}, combines such two forecasting methods in an efficient way as follows:
\begin{equation} \label{eq:PAR}
\hat{y}_d^{\rm \PAR}(t|a_1,...,a_n,b_0) =  a_1 \cdot {y}_d(t-1) + \cdots + a_n \cdot {y}_d(t-n) + b_0 \cdot \hat{y}_d^{\rm PM}(t).
\end{equation}
The model parameters $a_1$, $a_2$, ..., $a_n$, $b_0$ are to be trained. Furthermore ${\rm PM}$ here refers to the $N$-day model. The following setup is used $n=4$, $N=10.$

As an extension of the \PAR{} model we also introduce the \PARW{} model which also incorporates weather forecasts. The model retains the benefits of \PAR{}, namely the ability to capture daily and hourly patterns in the load and it also captures the weather impact, specially solar radiation feature selected according to the existence of an impact on the community consumption of PV panels installed on buildings, as follows: 
\begin{eqnarray}    \label{eq:PARW}
\lefteqn{\hat{y}_d^{\rm PARW}(t|a_1,...,a_n,b_0,c_0) =} \cr &&  a_1 \cdot {y}_d(t-1) + \cdots + a_n \cdot {y}_d(t-n) + b_0 \cdot \hat{y}_d^{\rm PM}(t) + c_0 \cdot \hat{y}_s(t),
\end{eqnarray}
where $\hat{y}_s(t)$ denotes the estimate of solar radiation of day $d$ at time interval $t$.

\subsection{Seasonal persistence-based regressive (\SPR{}) model}  \label{sec:SPRmodel}

The \SPR{} model, introduced in \cite{kychkin_feature_2021}, explores causal dependencies that are specific to the user-behavior in daily patterns of the electricity load, such as the average load over an extended window of time, or the total energy consumption within a day. For example, when the average total energy consumption within the morning hours has already been observed, this implies that the usually scheduled morning activities have already taken place, and therefore the expected load consumption in the upcoming hours will be low. More specifically, the \SPR{} model is defined using linear regression as follows: 
% \begin{eqnarray}    \label{eq:SPRModel}
% \lefteqn{\hat{y}_d^{\rm \SPR}(t|a_0,a_1,...,a_n) =  }\cr && a_0 \cdot f_d + a_1 \cdot y_{d-1}(t) + a_2 \cdot y_{d-7}(t) + a_3 \cdot y_{rs,d-1}(t) + a_4 \cdot y_{rs,d-7}(t) + \cr && 
% a_5 \cdot y_{h,d-1}(t) + a_6 \cdot y_{h,d-7}(t) + a_7 \cdot y_{d,d-1}(t) + a_8 \cdot y_{d,d-7}(t) + \cr &&
% a_9 \cdot y_{dh,d-1}(t) + a_{10} \cdot y_{dh,d-7}(t) + a_11 \cdot y_{{\rm low},d-1}(t) + a_{12} \cdot y_{{\rm low},d-7}(t) + \cr && 
% a_{13} \cdot y_{{\rm high},d-1}(t) + a_{14} \cdot y_{{\rm high},d-7}(t),
% \end{eqnarray}
\begin{eqnarray}    \label{eq:SPR}
\lefteqn{\hat{y}_d^{\rm \SPR}(t|a_0,a_1,...,a_n) = } \cr && a_0 \cdot f_d + \sum_{m=\{1,7\}} a_m y_{d-m}(t) + \sum_{\omega\in\Omega}\sum_{m=\{1,7\}} a_{\omega,m} \cdot y_{\omega,d-m}(t),
\end{eqnarray}
where $a_0$, $a_m$ and $a_{\omega,m}$ denote real coefficients and $N$ is the size of the history in days. The term $f_d$ denotes the categorical feature value that corresponds to the type of the day (working day or weekend), and $\Omega$ denotes the finite set of types of features that we use in the model where $\omega$ is a representative element of this set. In particular the features included in $\Omega$ include (i) the rolling-sum of the electricity load, (ii) the total electricity load within the current hour, (iii) the relative consumption between the current $15min$-interval load and the total consumption within day $d$, (iv) the difference in the hourly load within the last two hours, (v) a low energy consumption flag (less than 20\% of the daily mean load) and (vi) a high energy consumption flag (more than 150\% of the daily mean load). In principle, such aggregating features should better capture trends and causal dependencies in the consumption within one day. The possibilities of introducing additional features or adaptations is in fact open-ended.

\subsection{Seasonal persistence-based neural network (\SPNN) model} \label{sec:SPNNmodel}

The regression problem in Section~\ref{sec:SPRmodel} could also be solved in nonlinear manner by using artificial neural networks within the same feature set \cite{kychkin_feature_2021}. The resulting \SPNN{} model captures more complex dependencies between the features, which may reveal new patterns in users’ behavior, approximated by a \emph{Multi-layer Perceptron} (\MLP{}). \SPNN{} architecture provides a few layers, e.g. the \emph{input layer} consists of a number of neurons that coincides with the input features of the \SPR{} model (\ref{eq:SPR}); the \emph{hidden layer} includes four neurons and the \emph{output layer} consists of a single neuron, which provides the forecast of a given time instance. The \ReLU{} function is used as an \emph{activation function} and the mean squared error as a \emph{loss function}.

\section{Hourly-based \STLF{} models} \label{sec:HourlyModel}

In this section, we will introduce a collection of new models that can be utilized for the generation of hourly-based forecasts. As discussed in the introduction, such models could be particularly useful when making decisions regarding the use of energy flexibility in an hourly basis. Hourly forecasting models operate on a shorter horizon of a few hours ahead. Such models  run in a strongly scheduled way at specific time instances determined by the prediction horizon, e.g., when the prediction horizon is 4 hours, then predictions are updated at times 00:00, 04:00, 08:00, etc. assuming that the data for all previous measurements, e.g. 15 minutes ago, are known at the time of the update. 

\subsection{Adaptation of background forecasting models}

The adaptation of the forecasting methods presented in   Section~\ref{sec:ForecastingModelsBackground} to generating hourly forecasts could be straightforward for some of the models. However, for certain models such adaptations may not well reflect the design criteria of the original model, or they may lead to a higher or prohibitive computational complexity.

In particular, for certain classes of models, such as persistence models and auto-regressive-based models, the adaptation to hourly forecasts could be done in a rather straightforward manner. This is the case for the \SARIMA{} model, requiring only an adaptation to the frequency of generating forecasts (e.g., every 4 hours) and the use of the most recent available measurements. We will denote by \SARIMAH{} the hourly-based version of the original \SARIMA{}. Similar is also the case for the \HW{} model. The one-step ahead prediction provided by the model (when $k=1$) and the definition of the trend component that uses one-step load differences allow for generating forecasts in a more frequent basis utilizing the most recent available data. We will denote by \HWH{} the hourly-based version of the original \HW{} model.

The adaptation of the \PAR{} model is also rather straightforward and it will be discussed in detail in a forthcoming section. The corresponding version of the \SPR{} model requires additional modifications, which is one of the contributions of this paper, and will be discussed also in a forthcoming section. 

Contrary to the forecasting models that utilize auto-regressive features, the \Prophet{} and \LSTM{} models are designed to capture daily seasonal patterns and cannot be adapted to utilize most recent hourly measurements (given that there are no repeated patterns of the load consumption within one day). Therefore, forecasts may only be generated in a day-ahead fashion for these models. 

More specifically, the \Prophet{} model is determined by a daily seasonal term, a term capturing the effect of special days, and a trend term that depends on the position of the considered interval within the day. Thus, more recent measurements within one day cannot influence the upcoming forecasts for the upcoming hours within the same day. As a consequence, \Prophet{} is evaluated in the context of day-ahead forecasts only.  

\LSTM{} is also not appropriate for updating the forecasting sequences within one day. As was described in Section~\ref{sec:LSTM}, the \LSTM{} model receives the input data into blocks of sequences of $96$ daily measurements and provides a forecast as a sequence of length $96$. This setup is appropriate in order for the model to capture repeated daily patterns. In case it is necessary to update the forecasts more frequently within one day, e.g., every 4 hours, it would be necessary that different models are used for each one of the 4-hour blocks (that is 6 different models). In this case, each one of these models will be trained using blocks of historical sequences of length $16$ (of the corresponding 4 hour block and potentially also the previous 4 hour block). However, this separation of the model into 6 independent ones creates discontinuous forecasts, since each model tries to capture patterns that are observed within the corresponding 4 hour interval only and therefore it is not able to capture daily patterns that spread over the whole day. Secondly, the need for additional models (6 in the case of updating the forecasts every 4 hours or more in the case of more frequent updates) increases significantly and unnecessarily the computational complexity, including also the challenge for appropriately configuring its hyper-parameters. As a consequence, \LSTM{} is evaluated in the context of day-ahead forecasts only.  

\subsection{Hourly persistence-based auto-regressive models (\PARH{} and \PAReH{})}  \label{sec:PARHmodel}

As shown in Section \ref{sec:PARmodel}, the \PAR{} model combines auto-regressive and persistence factors. Transforming the model to provide hourly forecasts requires an adaptation of the auto-regressive features that need to utilize the most recently available measurements. In particular, the \PARH{} model takes on the following form:
\begin{equation}    \label{eq:PARH}
\hat{y}_{d,i}^{\rm \PARH}(t_i|a_1,...,a_n,b_0) =  a_1 \cdot {y}_{d,i}(t_i-1) + \cdots + a_n \cdot {y}_{d,i}(t_i-n) + b_0 \cdot \hat{y}_{d,i}^{\rm PM}(t_i).
\end{equation}
for some unknown parameters $a_1$, $a_2$, ..., $a_n$, $b_0$. Recall also the {\rm PM} estimate corresponds to the $N$-day model. An extended version is also considered, namely the Extended Hourly Persistence-based Auto-regressive \PAReH{} model, where estimates of both types of persistence models are considered, namely $N$-day and $N$-same-day, as additional features.
% \begin{eqnarray}    \label{eq:PAReH}
% \lefteqn{\hat{y}_{d,i}^{\rm PAReH}(t_i|a_1,...,a_n,b_0,c_0) = } \cr && a_1 \cdot {y}_{d,i}(t_i-1) + \cdots + a_n \cdot {y}_{d,i}(t_i-n) + b_0 \cdot \hat{y}_{d,i}^{\rm Nsameday}(t_i) + \cr &&  c_0 \cdot \hat{y}_{d,i}^{\rm Ndays}(t_i),
% \end{eqnarray}
% where $\hat{y}_{d,i}^{\rm Nsameday}(t_i)$ denotes the estimate of the load by N-same-day persistence model and $\hat{y}_{d,i}^{\rm Ndays}(t_i)$ denotes the estimate of the load by N-days persistence model at at time $t$ within $i$-th 4 hour interval of day $d$.

\subsection{Hourly seasonal persistence-based regressive (\SPRH{}) model}  \label{sec:SPRH}

The \SPR{} model introduced in Section~\ref{sec:SPRmodel} can  capture causal dependencies in the use of electricity. In the case of hourly forecasts these dependencies may be more influential and therefore improve the upcoming forecasts. This is the motivation of introducing the corresponding hourly version of the \SPR{} model. In particular, we introduce the Hourly Seasonal Persistence-based Regressive model, denoted briefly by \SPRH{}, as follows:
% \begin{eqnarray}    \label{eq:SPRHModel}
% \lefteqn{\hat{y}_d^{\rm \SPRH}(t|a_0,a_1,...,a_m) = a_0 \cdot f_d + a_1 \cdot f_{d-1} + }\cr && 
% a_2 \cdot y_{d,l}(t) + a_3 \cdot y_{d-1,n}(t) + a_4 \cdot y_{d-1}(t) + a_5 \cdot \frac{1}{N}\sum_{i=d-2-N}^{d-2}y_{i}(t) + \cr && 
% a_6 \cdot y_{rs,d,l}(t) + a_7 \cdot y_{rs,d-1,n}(t) + a_8 \cdot y_{rs,d-1}(t) + a_9 \cdot \frac{1}{N}\sum_{i=d-2-N}^{d-2}y_{rs,i}(t) +\cr && 
% a_{10} \cdot y_{h,d,l}(t) + a_{11} \cdot y_{h,d-1,n}(t) + a_{12} \cdot y_{h,d-1}(t) + a_{13} \cdot \frac{1}{N}\sum_{i=d-2-N}^{d-2}y_{h,i}(t) +\cr &&
% a_{14} \cdot y_{dh,d,l}(t) + a_{15} \cdot y_{dh,d-1,n}(t) + a_{16} \cdot y_{dh,d-1}(t) + a_{17} \cdot \frac{1}{N}\sum_{i=d-2-N}^{d-2}y_{dh,i}(t) +\cr &&
% a_{18} \cdot y_{p,d,l}(t) + a_{19} \cdot y_{p,d-1,n}(t) + a_{20} \cdot y_{p,d-1}(t) + a_{21} \cdot \frac{1}{N}\sum_{i=d-2-N}^{d-2}y_{p,i}(t) +\cr &&
% a_{22} \cdot y_{low,d,l}(t) + a_{23} \cdot y_{low,d-1,n}(t) + a_{24} \cdot y_{low,d-1}(t) + a_{25} \cdot \frac{1}{N}\sum_{i=d-2-N}^{d-2}y_{low,i}(t) +\cr &&
% a_{28} \cdot y_{high,d,l}(t) + a_{29} \cdot y_{high,d-1,n}(t) + a_{30} \cdot y_{high,d-1}(t) + \cr &&
% a_{31} \cdot \frac{1}{N}\sum_{i=d-2-N}^{d-2}y_{high,i}(t),
% \end{eqnarray}
\begin{eqnarray}    \label{eq:SPRH}
\lefteqn{\hat{y}_d^{\SPRH}(t|a_0,a_1,...,a_n) = } \cr && a_0 \cdot f_d + \sum_{m=1}^{N} a_m y_{d-m}(t) +  \sum_{\omega\in\Omega}\sum_{m=1}^{N} a_{\omega,m}  y_{\omega,d-m}(t) + \cr && \sum_{\omega\in\Omega} a_{\omega,-1} y_{\omega,d}(t-\tau_{-1}(t)) + \sum_{\omega\in\Omega} a_{\omega,+1} y_{\omega,d-1}(t-\tau_{+1}(t)),
\end{eqnarray}
where $a_0$, $a_m$, $a_{\omega,m}$, $a_{\omega,-1}$ and $a_{\omega,+1}$ denote real coefficients, $N$ is the size of history in days, $\tau_{-1}(t)\df \lfloor \nicefrac{t}{K_h} \rfloor\cdot t$ is the time index of the previous forecast update within the day, and $\tau_{+1}(t)\df \lfloor \nicefrac{t}{K_h} \rfloor\cdot t$ is the time index of the next forecast update. Furthermore, $f_d$ denotes the categorical feature value that corresponds to the type of the day (working day or weekend). The finite set $\Omega$ is the collection of types of features that we use in the model where $\omega$ is a representative element of this set. The set $\Omega$ is defined as in the case of the \SPR{} model of Section~\ref{sec:SPRmodel}. 

The first three terms of the model coincide with the corresponding ones of the day-ahead version of the model in Section~\ref{sec:SPRmodel}. These terms capture the effect of the type of the day, and also persistence factors over a history of $N$ days. The last two terms create an estimate of the features in $\Omega$, based on the last measured feature values $y_{\omega,d}(t-\tau_{-1}(t))$, given that the load measurements at time $\tau_{-1}(t)$ are available, and the next estimated feature values $y_{\omega,d-1}(t-\tau_{+1}(t))$ using the measurements of the previous day. Given that the features included in $\Omega$ entail averaging quantities (e.g., rolling sum) or differences, including the most recent estimates for both $\tau_{-1}(t)$ and $\tau_{+1}(t)$ provides a better smoothing of the estimates.

\section{Evaluation}
\label{sec:Evaluation}

In this section, we perform a comparative analysis for both day-ahead and hourly forecasts. Data from a small energy community of three residential buildings are utilized from Wels, Upper Austria. The available data are utilized in an iterative or simulation manner, where at the end of each day, all models are trained using all available previous data. The trained models are then used to generate forecasts over the next day, or next hours. The root mean square error (\RMSE{}) is then calculated between the forecasts and actual load measurements.

First, we form a baseline forecast using the persistence $N$-day and $N$-same-day models. For a more thorough benchmark testing we also use \XGBoost{} to generate ensembles of variations of persistence models. Day-ahead forecasts are also generated by all considered models to better highlight the main differences and capabilities of the models (given also the fact that certain models can only be applied in a daily basis such as the \Prophet{} and \LSTM{} models). On the other hand, the models that incorporate auto-regressive features can be used for hourly forecasts, including \HWH{}, \SARIMAH{}, and the newly introduced \PARH{}, \PAReH{}, \SPRH{}. We evaluated the performance of these models in terms of the relative \RMSE{} and over several \emph{simulation/test periods}, lasting from $1$-month for different months to $1$-year. 

\begin{figure}
    \centering
    \includegraphics[width=1.\linewidth]{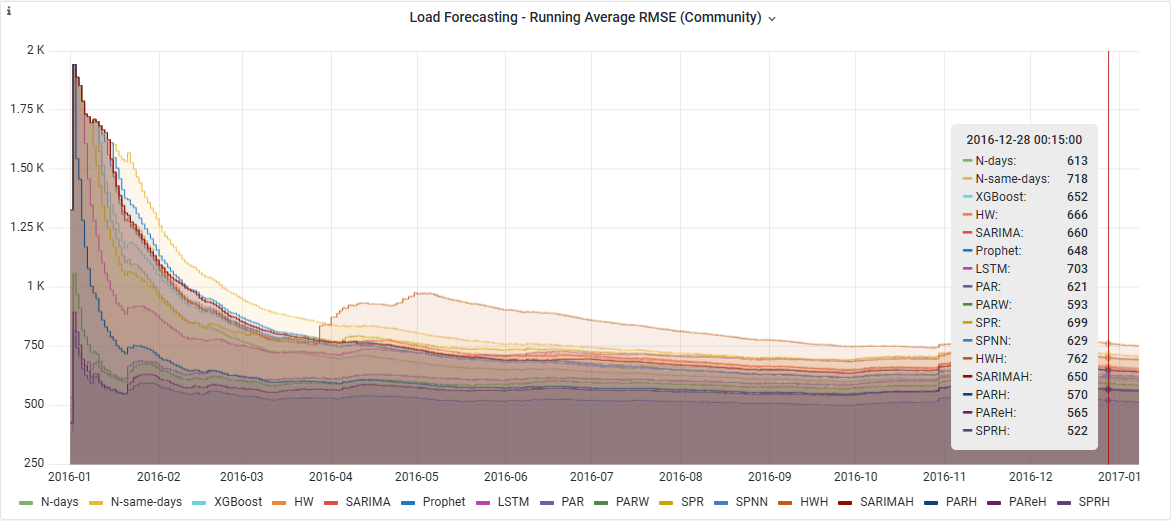}
    \caption{Running average \RMSE{} of 4-hour ahead predictions in a small energy community of three residential buildings in Wels, Upper Austria.}
    \label{fig:RunningAverageRMSE}
\end{figure}

Table~\ref{tb:ResultsrRMSE} and Figure~\ref{fig:RunningAverageRMSE} provide a comparative analysis of the models with respect to prediction accuracy both in daily and hourly basis. Several observations have been noted. First, persistence models, such as the $N$-day model, can provide rather robust and accurate forecasts throughout the testing period, and can outperform powerful models such as \XGBoost{}, \SARIMA{}, \Prophet{}, \LSTM{}. Furthermore, models that combine both persistence and auto-regressive features can significantly improve over the standard persistence models, see, e.g., the \PAR{} significantly improves accuracy over the $N$-day model. Although \SPR{} does include persistence factors, it is not able to improve over \PAR{}, possibly because it only uses a limited number of persistence factors (the previous day, and the previous-same-day), compared to the $N$-day model that uses all $10$ previous days. Furthermore, the causal effects modeled by the \SPR{} model may only have an impact under sub-day measurement updates (as done in the hourly forecasts).  This becomes evident indeed when the models are compared in the hourly forecasts, where \SPRH{} outperforms both \PARH{} and \PAReH{}.

\begin{table}[h!]
\begin{center}
\caption{Relative average RMSE for an energy community over the average electricity load in 2016. The first group of models are evaluated with respect to day-ahead forecasts, while the second group of models are evaluated with respect to hourly forecasts with an update period and prediction horizon of $4$ hours. } \label{tb:ResultsrRMSE}
\begin{tabular}{cccccccc}
Model & Jan. & Feb. & June & July & Nov. & Dec. & One year  \\\hline\hline 
N-day       & 0.991 & 0.664 & 0.766 & 0.752 & 0.379 & 0.467 & 0.621 \\
N-same-day  & 1.170 & 0.843 & 0.889 & 0.868 & 0.437 & 0.546 & 0.731 \\
\XGBoost{}	& 1.020 & 0.715	& 0.811 & 0.795	& 0.406 & 0.497 & 0.659 \\
\HW{}	    & 0.936 & 0.713	& 0.854 & 0.846	& 0.427 & 0.527 & 0.678 \\
\SARIMA{}   & 0.904 & 0.632	& 0.806 & 0.801	& 0.405 & 0.497 & 0.634 \\
\Prophet{}  & 1.110 & 0.766	& 0.809 & 0.768 & 0.392 & 0.490 & 0.667 \\
\LSTM{}     & 0.837 & 0.616	& 0.849 & 0.843 & 0.435 & 0.535 & 0.653 \\
\PAR{}      & \textbf{0.496} & 0.500 & 0.726 & 0.724 & 0.380 & 0.474 & 0.543 \\
\PARW{}     & 0.523 & \textbf{0.489} & \textbf{0.711} & \textbf{0.711} & \textbf{0.370} & \textbf{0.455} & \textbf{0.532} \\
\SPR{}      & 0.952 & 0.669 & 0.867 & 0.852 & 0.435 & 0.533 & 0.682 \\
\SPNN{}     & 1.130 & 0.789 & 0.827 & 0.803 & 0.396 & 0.487 & 0.681 \\ \hline 
\HWH{}      & 1.080 & 0.730 & 1.040 & 0.999 & 0.471 & 0.580 & 0.773 \\
\SARIMAH{}  & 1.100 & 0.757 & 0.818 & 0.816 & 0.403 & 0.494 & 0.671 \\
\PARH{}	    & 0.674 & 0.509	& 0.680 & 0.676	& 0.349 & 0.432 & 0.529 \\
\PAReH{}	& 0.454 & 0.455	& 0.668 & 0.664	& 0.347 & 0.429 & 0.498 \\
\SPRH{}	    & \textbf{0.440} & \textbf{0.433} & \textbf{0.617} & \textbf{0.615} & \textbf{0.321} & \textbf{0.398} & \textbf{0.464} \\\hline

\end{tabular}
\end{center}
\end{table}

\section{Conclusions and Future Work}
\label{sec:Conclusions}

In this paper, different types of predictive models were utilized for generating day-ahead and hourly-based forecasts. Models ranged from standard persistence-models to more technologically advanced, e.g., based on neural networks. From the presented analysis, it has become clear that models of simple structure, such as the $N$-day model, or models that combine persistence and auto-regressive features can outperform more advanced and computationally intensive models, such as the \SARIMA{} and \LSTM{} models. Furthermore, when extending the investigation to hourly-based forecasts, it has also become clear that not all models are appropriate, either due to their design principles (as in the case of the \Prophet{} model) or due to their computational complexity (as in the case of the \LSTM{} model). The comparative analysis in hourly forecasts has also revealed that computationally efficient models (based on linear regression), such as the introduced \SPRH{}, can outperform more advanced techniques (such as \SARIMAH{}) by better exploiting the causal effects present in electricity load consumption. This observation also drives our future work where causality dependencies should further be explored in the generation of hourly and daily forecasts.

% \begin{ack}
% \end{ack}

%% The Appendices part is started with the command \appendix;
%% appendix sections are then done as normal sections
%% \appendix

%% \section{}
%% \label{}

%% If you have bibdatabase file and want bibtex to generate the
%% bibitems, please use
%%
% \section{Bibliography}  \label{sec:Bibliography}

\bibliographystyle{ieeetr} 
\bibliography{bibliography}

%% else use the following coding to input the bibitems directly in the
%% TeX file.

% \begin{figure}
%     \centering
%     \includegraphics[scale=0.2]
%     {./STLF_MindMap_new2.png}
%     \caption{MindMap of buildings load forecasting domain}
%     \label{fig:MindMap}
% \end{figure}

\end{document}
\endinput
%%
%% End of file `elsarticle-template-num.tex'.